\documentclass[conference,comsoc]{IEEEtran}

\usepackage{cite,graphicx,amsmath,amssymb}
\usepackage{amsfonts,amssymb}
\usepackage{subfigure}
\usepackage{fancyhdr}
\usepackage{mdwmath}
\usepackage{mdwtab}
\usepackage{balance}
\usepackage{xcolor}
\usepackage{bm}
\usepackage{amsthm}
\usepackage{algorithm}
\usepackage{algorithmic}
\usepackage{multirow}
\usepackage{flafter}
\usepackage{mathrsfs}
\usepackage{url}
\usepackage{multicol}

\theoremstyle{definition}

\newtheorem{corollary}{Corollary}

\usepackage[left= 0.625in, right= 0.625in,top=0.76in, bottom=0.97in]{geometry}

\begin{document}

\title{Recursive Euclidean Distance Based Robust Aggregation Technique For Federated Learning}

\author{
{Charuka Herath,  Yogachandran Rahulamathavan , Xiaolan Liu} \\
\IEEEauthorblockA{Institute for Digital Technologies, Loughborough University}

Emails: \{c.herath, y.rahulamathavan, xiaolan.liu\}@lboro.ac.uk
}

\maketitle

\begin{abstract}

Federated learning has gained popularity as a solution to data availability and privacy challenges in machine learning. However, the aggregation process of local model updates to obtain a global model in federated learning is susceptible to malicious attacks, such as backdoor poisoning, label-flipping, and membership inference. Malicious users aim to sabotage the collaborative learning process by training the local model with malicious data. In this paper, we propose a novel robust aggregation approach based on recursive Euclidean distance calculation. Our approach measures the distance of the local models from the previous global model and assigns weights accordingly. Local models far away from the global model are assigned smaller weights to minimize the data poisoning effect during aggregation. Our experiments demonstrate that the proposed algorithm outperforms state-of-the-art algorithms by at least $5\%$ in accuracy while reducing time complexity by less than $55\%$. Our contribution is significant as it addresses the critical issue of malicious attacks in federated learning while improving the accuracy of the global model.

\end{abstract}


\section{Introduction}
The emerging Artificial Intelligence market is accompanied by an unprecedented growth of cloud-based AI solutions. This technological revolution was catalyzed by rapidly expanding personal computing devices. Most people frequently carry their intelligent devices equipped with multiple sensors. As a result, personal computing devices offer access to a large amount of training data necessary to build reliable machine learning (ML) models. However, traditional ML requires gathering the training data in a single machine or a data center. As a result, technology companies must go through the costly and lengthy process of harvesting their users’ data, not mentioning the risks and responsibilities of storing data in a centralized location. This also leads to many privacy violation issues.\footnote{\textcolor{red}{This paper has been submitted to IEEE IAS GlobeNet 2023 Conference.}}

Federated Learning (FL) is developed to address these issues by enabling end-user devices to collaboratively learn a shared global model using the locally-stored training data under the orchestration of a central server, decoupling training deep learning models from the need to collect and store the data in the cloud. With its decentralized data approach, FL is one of the fastest-growing research fields.

\begin{figure}[t]
  \centering
  \includegraphics[width=3.4in]{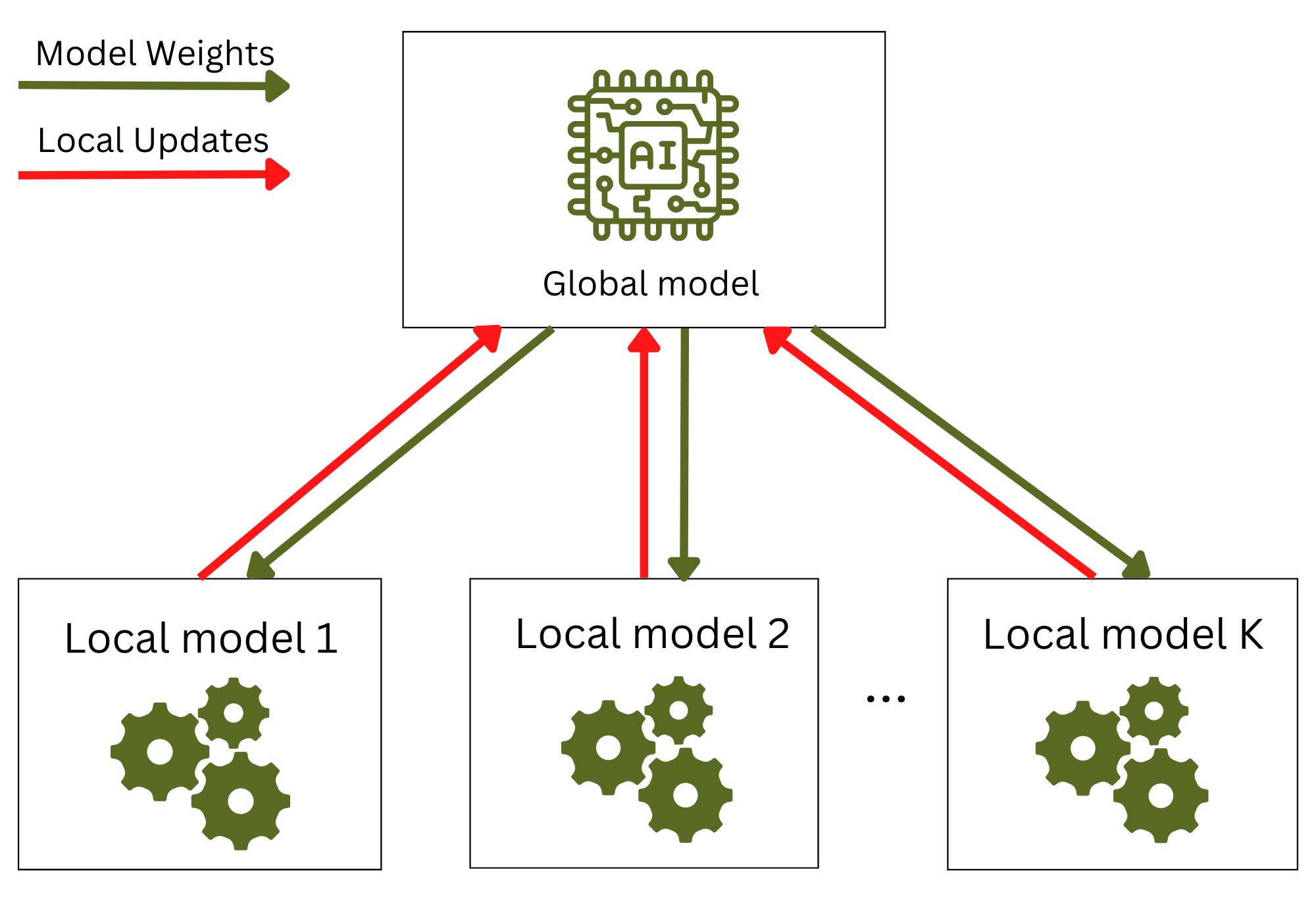}\\
  \caption{Federated Learning Framework}\label{Federated_Learning_Framework}
\end{figure}

Generally, the client and the server are the two prominent roles in FL. The client is the owner of training data, and the server is the aggregator of the global model parameters. FL is characterized by obtaining better global model performance while keeping all client-side training data. As shown in Fig. \ref{Federated_Learning_Framework}, In the FL architecture, the server initializes the parameters of the global model and distributes them to the clients participating in FL. Secondly, the clients train their local models using their data. Thirdly, they upload the new parameters of their models to the server for aggregation. The above process is repeated until the pre-defined condition is met \cite{Junchuan2022}.

By considering the distribution of the data, data can be mainly divided into two parts, which are identically distributed data (IID) and non-independently and identically distributed data (Non-IID).

\begin{figure*}
\centering
\hspace{-15mm}
\subfigure[Horizontal FL]{
\includegraphics[scale=0.28]{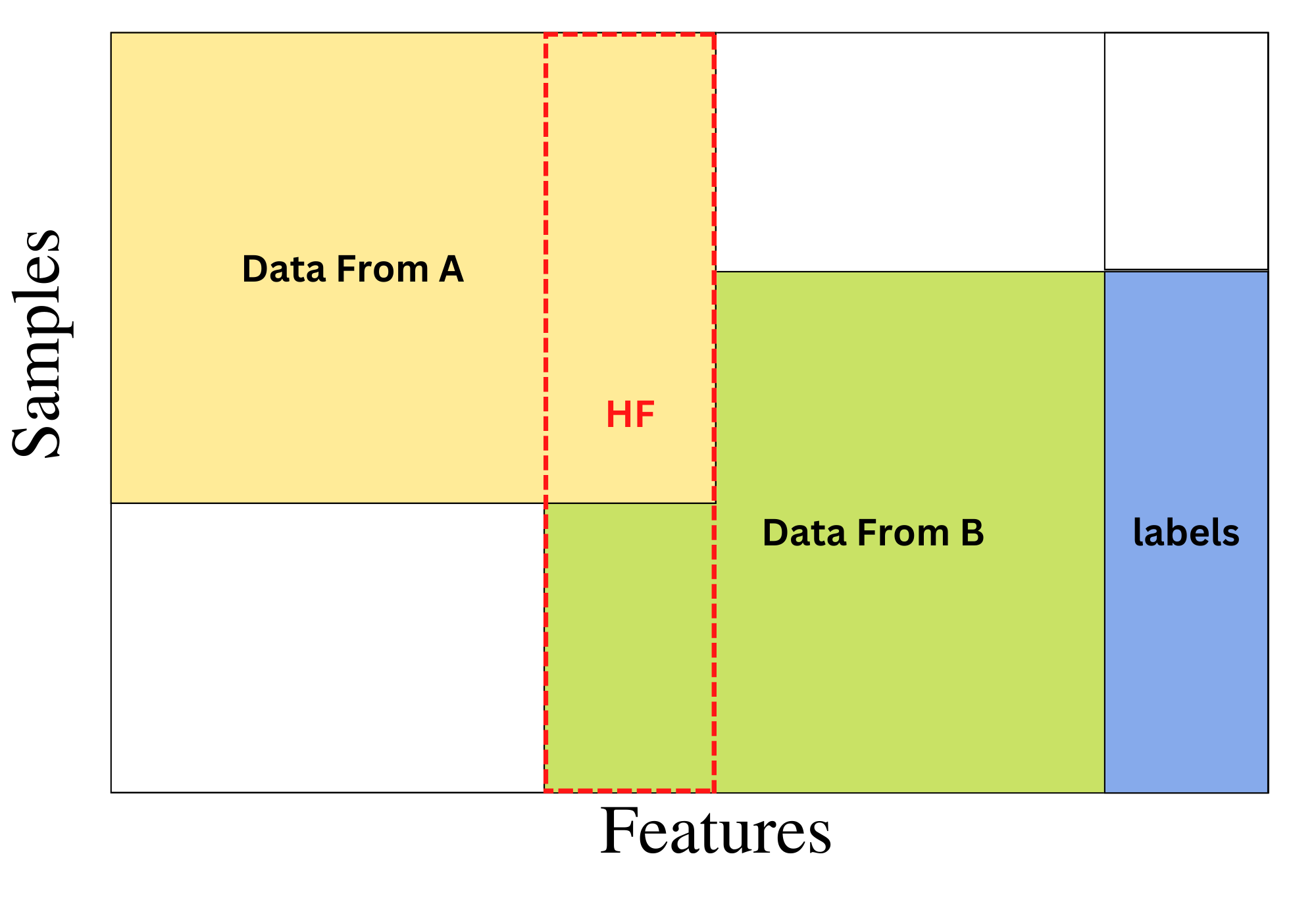}
\label{horizontal_fL}
}\hspace{-2.5mm}
\subfigure[Vertical FL]{
\label{vertical_fL}
\includegraphics[scale=0.28]{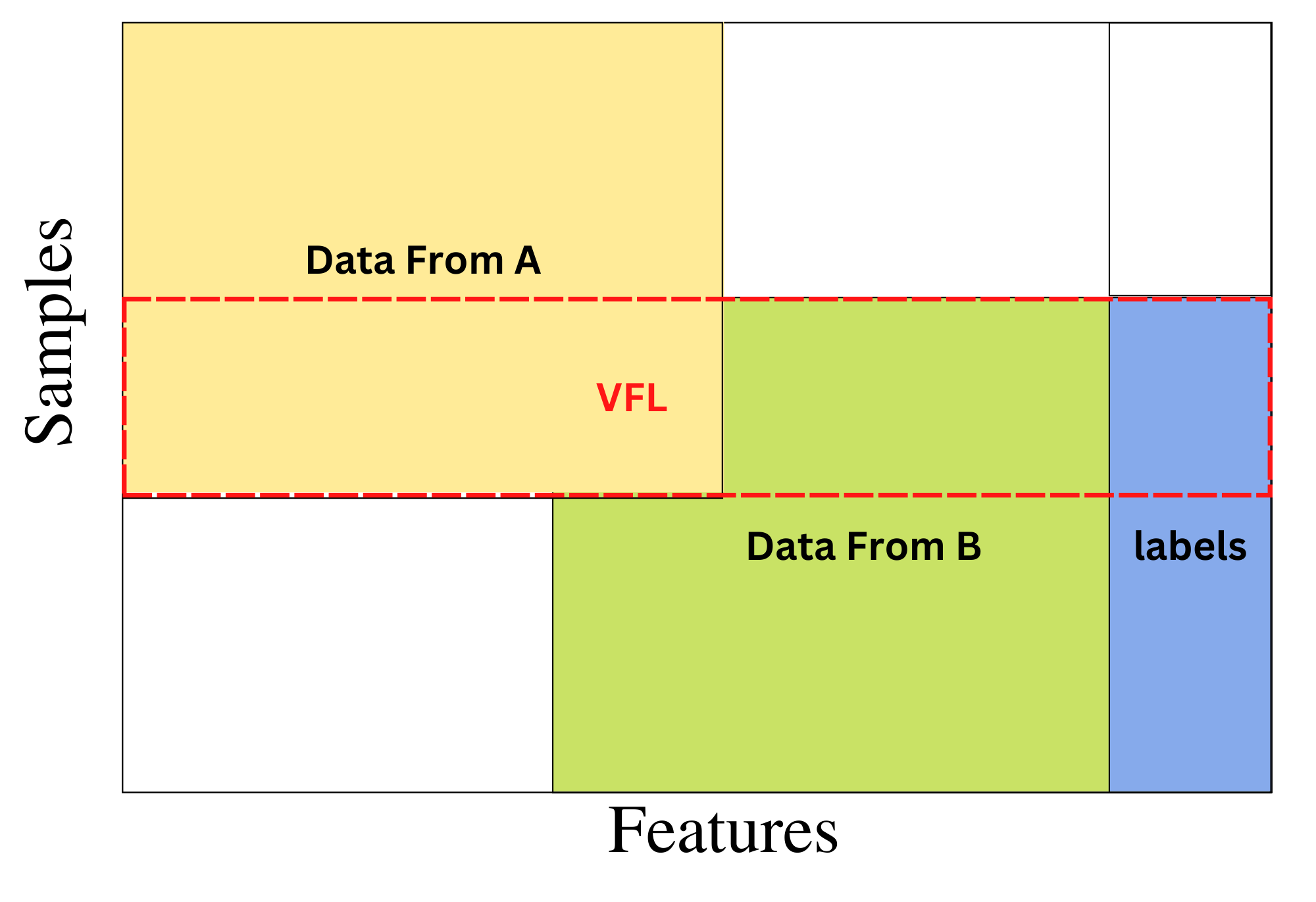}
}\hspace{-2.5mm}
\subfigure[Transfer FL]{
\label{Federated_transfer_learning}
\includegraphics[scale=0.28]{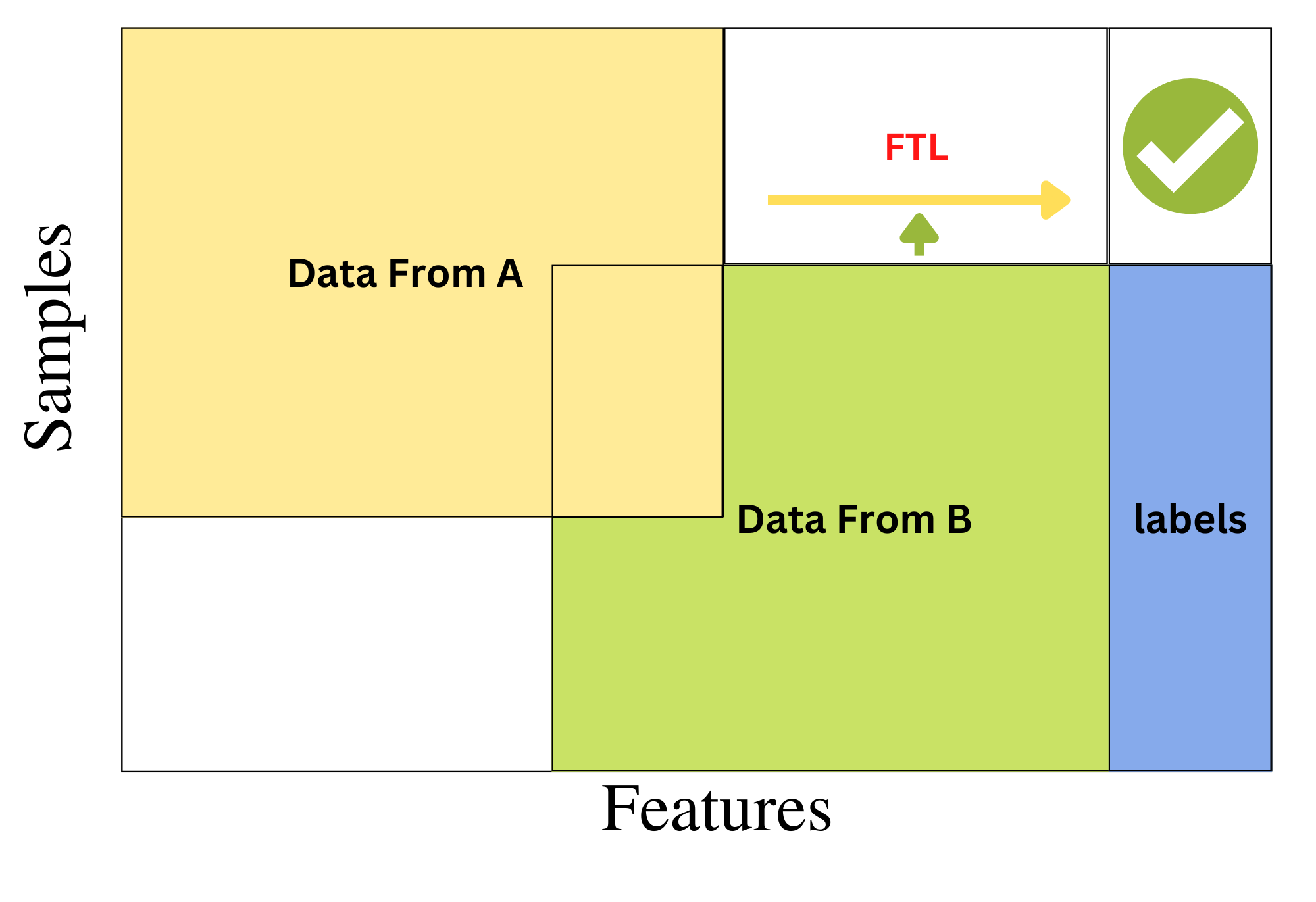}
}\hspace{-10.5mm}

\caption{Data Distribution Methods in FL}
\label{data_distribution_FL}
\end{figure*}

Moreover, as stated in \cite{Qiang2019}, by considering the distribution of training data, FL can be categorized into the three groups shown in Fig. \ref{data_distribution_FL} which are horizontal FL, vertical FL, and transfer FL. 

\begin{itemize}
\item Horizontal FL: Horizontal FL or sample-based FL illustrated in Fig. \ref{horizontal_fL}, is introduced in the scenarios that data sets that share the same feature space but are different in samples. For instance, where, the same database is shared among two different hospitals that will share the same set of features, but different data samples from two different regions can be taken as an example.
\item Vertical FL:  Vertical FL or feature-based FL illustrated in Fig. \ref{vertical_fL}, applies to cases where two data sets share the same sample ID space but differ in feature space. For instance, two hospitals that treat dental issues (feature) and general issues (feature) maintain two databases for the same set of patients (samples) in one area can be taken as an example.
\item Transfer FL: Transfer FL illustrated in Fig. \ref{Federated_transfer_learning} applies to the scenarios that the two data sets differ not only in samples but also in feature space. For instance, training a personalized model for a movie recommendation for the user's past browsing behavior can be taken as an example.
\end{itemize}

In our study, we are focusing on HFL. However, future works will consider the other two data distribution architectures.

In a real-world scenario, data used for FL may contain biases, such as wrongly annotated labels, and missing data. Also, adversarial clients may sabotage the learning process by sending corrupted local models to the aggregation process. Standard aggregation methods such as Federated Averaging (FedAvg) purposed by \cite{Brendan2017} are vulnerable to these bad local models. Moreover, \cite{Shuhao2019} states that FL suffers from label-flipping and backdoor attacks. When a local model is poisoned, the aggregated global model can also be poisoned and fail to behave correctly.

To tackle these challenges, researchers developed many defense mechanisms, such as robust aggregation, zero-knowledge proof, and recognizing legitimate clients \cite{Nader2021}. In this study, we develop novel robust aggregation mechanisms for horizontally distributed data under an IID setting to make FL more attack-resistance against label flipping attacks. Also, We compare our proposed algorithm with \cite{Shuhao2019} purposed Residual-based Reweighting aggregation algorithm and the standard averaging algorithm by conducting experiments on the MNIST dataset. Our proposed aggregation algorithm significantly mitigates the impact of label-flipping  models on the IID setting and outperforms other baselines in terms of the lesser average attack success rate of 1.1962\%, higher average model accuracy of 98.27\%, and lesser average aggregation time of 0.028s. 

\section{problem Formulation}

A study in \cite{Reza2016} proposed an FL method based on sampling a fraction of the gradients for every client, known as Federated Stochastic Gradient Decent (FedSGD). The gradients are then sent to a global server and averaged proportionally to the number of samples on each client.

FedAvg by \cite{Brendan2017} illustrated in the \textbf{Algorithm \ref{Fed_avg_algo}} is a more efficient and generalized version of FedSGD by allowing clients to train their local models on multiple batches of local data. Then, the parameters of the model are shared with the global server instead of the gradients as in FedSGD. Hence, FedAvg can be defined as the weighted average of the local updated model parameters ${M}_n^k$ sent by the ${K}$ clients. Here ${n}$ is the training round out of ${N}$ rounds.
\begin{equation}\label{fedAvg}
{M_n} = \sum\limits_{k=1}^K {\frac{D_k}{D}{M}_n^k}
\end{equation}

\begin{algorithm}
\caption{Standard Federated Averaging Algorithm}
\label{Fed_avg_algo}
\begin{algorithmic}[1]
\renewcommand{\algorithmicrequire}{\textbf{Initialization:}}
\REQUIRE~\\
\textbf{Input:} Locally trained models ${M_n^1}$, ${M_n^2}$,..., ${M_n^k}$ from each participants for the $n^{th}$ epoch\\
\renewcommand{\algorithmicensure}{\textbf{Learning:}}
\ENSURE~\\
\FOR {$k \leq K $}
\STATE Add each local model together while averaging by partition data used by ${k^{th}}$ client
\ENDFOR
\STATE Average the sum of models \\
${M_n} = \sum\limits_{k=1}^n {\frac{D_k}{D}{M}_n^k}$
\STATE \textbf{Return } ${M_n}$.
\end{algorithmic}
\end{algorithm}

The FL method has some critical issues regarding the approach feasibility and, most importantly, privacy and security. Therefore, in our research, we are focusing on privacy and security-related issues. As pointed out in the introduction section, poisoning attacks are the most critical issue in both FL and ML. An attack injects vulnerabilities by a malicious attacker to manipulate the global model. Poisoning attacks can be divided into two types, i.e., targeted poisoning attacks and untargeted poisoning attacks \cite{Nader2021}. 

A targeted poisoning attack, i.e., model poisoning, aims to completely control or affect one sub-task of the FL model without affecting any other sub-task. In a backdoor attack, the global model behaves incorrectly on adversarial targeted input [3]. For instance, for an image classification application, the attacker may corrupt the model to misclassify "red cars" as "bikes" while ensuring other car instances are classified correctly.

The purpose of untargeted poisoning attacks, i.e., label flipping, is to corrupt or manipulate an FL model by changing the model's training data or parameters, which allows the attacker to control some parameters of the training process of some selected clients, such as raw data, training rounds, and model parameters. A study by \cite{Matei2020} shows that the label-flipping attack has great harm to a federated system even with a minimal number of attackers. In this attack, the attacker flips the training data labels from one class to another and trains the model accordingly.

However, the default FL aggregation algorithm is not immune to attacks and failures that target each step of the system’s training and deployment pipelines. \cite{Yuzhe2019},\cite{Nader2021}, \cite{Arjun2019}, \cite{Eugene2018}, \cite{Ruikang2022}, show that under the traditional aggregation approach, Fl suffers from different attacks such as label flipping and backdoor attacks. When a local model is poisoned, the aggregated global model can also be poisoned and fail to
behave correctly.


\section{Background and Related Works}

In the broader picture, data anomalies are vital in centralized and decentralized machine-learning approaches. So, ensuring the trustworthiness of data is critical. Developing a novel approach for secure FL will solve data availability and privacy issues.

\subsection{Robust Aggregation Mechanisms}

Aggregation algorithms security is elemental to secure FL. There is a breadth of research on robust aggregation algorithms in \cite{Matei2020}, that detect and discard faulty or malicious updates during training. However, \cite{Peter2019} states that many state-of-the-art robust aggregations rely on realistic assumptions or need to be held in FL environments. As a result, several novel aggregation methods have been proposed, such as adaptive aggregation. This aggregation method is robust against model update corruption in up to half the clients. Furthermore, \cite{Yanyang2020} suggested using Gaussian distribution to measure clients’ potential contributions. While this method is effective, evaluating each model in every round requires much time.  Statistical methods have been studied and applied in robust distributed learning where data is IID. For example, the median and trimmed mean methods are practical approaches in robust distributed learning.

One of the novel robust aggregation mechanisms purposed by \cite{Shuhao2019}, combines repeated median regression using residual distance with a reweighting scheme that is iteratively reweighted least squares (IRLS). While this method is effective, it requires comparatively high time to reweight each local model parameter. Also, it accumulates the parameter confidence in each local model in every single round or training. Even though this method is one of the successful attack-resistant robust aggregation methods, the time complexity of the aggregation algorithm is ${O(n)}$. Test results show that our aggregation reduces the aggregation time by 50\% while maintaining a high attack-resistant rate and a high model accuracy.

\section{Proposed Solution}

\subsection{Euclidean Distance}
Euclidean distance is a widely used distance metric. It is used in many ML algorithms as a default distance metric to measure the similarity between two recorded observations. Moreover, it works on the principle of the Pythagoras theorem and signifies the shortest distance between two points illustrated in (\ref{eucDist}).

\begin{equation}\label{eucDist}
{E(x,y)} = \sqrt{\sum\limits_{i=1}^n {(x_i - y_i)^2}}
\end{equation}

\textbf{Algorithm \ref{Euc_dist_algo}} summarizes our aggregation algorithm, and a detailed step-by-step description is provided below. Here, in each epoch, we are adjusting the weights of each local participant by using its Euclidean distance to the most recent global model weights during the aggregation process.

\subsection{Aggregation Algorithm}
\textbf{Model initialization.} There are multiple rounds of communication between participants and a central server for learning a global model in FL. In each round, the global model is shared among the \textit{K} participants. First, a local model on each device is trained on its local private data with the shared global model as initialization. Then all the \textit{K} local models are sent to the central server to update the global model using the aggregation algorithm for \textit{N} rounds. For instance, let's denote the model trained by the ${k^{th}}$ user in the ${n^{th}}$ training round as \textit{$M_n^k$}. 

\textbf{Flatten local and global model weights into separate single-dimension arrays.} Here, we are feeding an array of local models and the current global model into the aggregation algorithm. Then we are flattening each model's layer into a single-dimension tensor illustrated in (\ref{single_dimention_tensor}). Here, ${t}$ is the number of weights in each model.

\begin{equation}\label{single_dimention_tensor}
{M_n} = {[x_n^1, x_n^2, ..., x_n^t]}, 
{M_n^k} = {[x_n^{k^1}, x_n^{k^2}, ..., x_n^{k^t}]} 
\end{equation}

\textbf{Calculation Euclidean Distance.} Then, at the central server, server, we calculate the Euclidean distance ${e^k}$ for each participant's local model with respect to the current global model ${M_n}$ during each epoch as illustrated in (\ref{euq_dist_calculation}).

\begin{equation}\label{euq_dist_calculation}
{e_k} = {|M_n - M_n^k|}
\end{equation}

\textbf{Average local model update.} Now, we average the local model update in proportion to its Euclidean distance as illustrated in (\ref{average_local_model}).

\begin{equation}\label{average_local_model}
{M_k^n} = {\frac{1}{e_k}{M_n^k}} 
\end{equation}

\textbf{Global model aggregation.} Finally, all local participant models will be aggregated into the global model illustrated in (\ref{aggregation}).
\begin{equation}\label{aggregation}
{M_n} = \frac{{\sum\limits_{k=1}^K{\frac{1}{e_k}{M_n^k}}}}{{\sum\limits_{k=1}^K{\frac{1}{e_k}}}}
\end{equation}

\begin{algorithm}[t]
\caption{Euclidean Distance-based Averaging Algorithm}
\label{Euc_dist_algo}
\begin{algorithmic}[2]
\renewcommand{\algorithmicrequire}{\textbf{Initialization:}}
\REQUIRE~\\
\textbf{Input:}  Locally trained models ${M_n^1}$, ${M_n^2}$,..., ${M_n^k}$ from each participants and the current global model ${M_n}$ for the $n^{th}$ epoch.\\
\renewcommand{\algorithmicensure}{\textbf{Learning:}}
\ENSURE~\\
\STATE Flatten local and global model.\\
${M_n} = {[x_n^1, x_n^2, ..., x_n^t]}$
\FOR {$k \leq K $}
\STATE Flatten ${k_{th}}$ local and global model\\
${M_n^k} = {[x_n^{k^1}, x_n^{k^2}, ..., x_n^{k^t}]}$
\STATE Calculation Euclidean distance for the ${k^{th}}$ participant.\\
${e_k} = {|M_n - M_n^k|}$
\STATE Average the ${k^{th}}$ participant's model.\\
${M_k^n} = {\frac{1}{e_k}{M_n^k}}$
\ENDFOR
\STATE Global model aggregation. \\
${M_n} = \frac{{\sum\limits_{k=1}^K{M_n^k}}}{{\sum\limits_{k=1}^K{\frac{1}{e_k}}}}$
\STATE \textbf{Return } ${M_n}$.
\end{algorithmic}
\end{algorithm}

\section{Simulation Results}
In this section, we conduct image recognition experiments on MINST dataset to illustrate the learning performance of our developed robust aggregation mechanism. We compare our approach FedAvg \cite{Brendan2017} and Residual-based Reweighting algorithm \cite{Shuhao2019}. We
perform experiments on the MNIST handwritten digit dataset and we implement attack strategies and defense algorithms in PyTorch. We use a two-layer convolutional neural network (CNN) for our MNIST experiments. With this simple CNN model, our goal is to evaluate different aggregation algorithms for defending FL in the presence of attacks.

\subsection{MNIST Dataset}
MNIST dataset. The MNIST dataset contains 70,000 real-world handwritten images with digits from 0 to 9. We evaluate different methods by learning a global model on these training images distributed on multiple devices in an IID setting with adversarial attacks.

\subsection{Results without any Attackers}
\begin{figure}[t]
  \centering
  \includegraphics[width=3.4in]{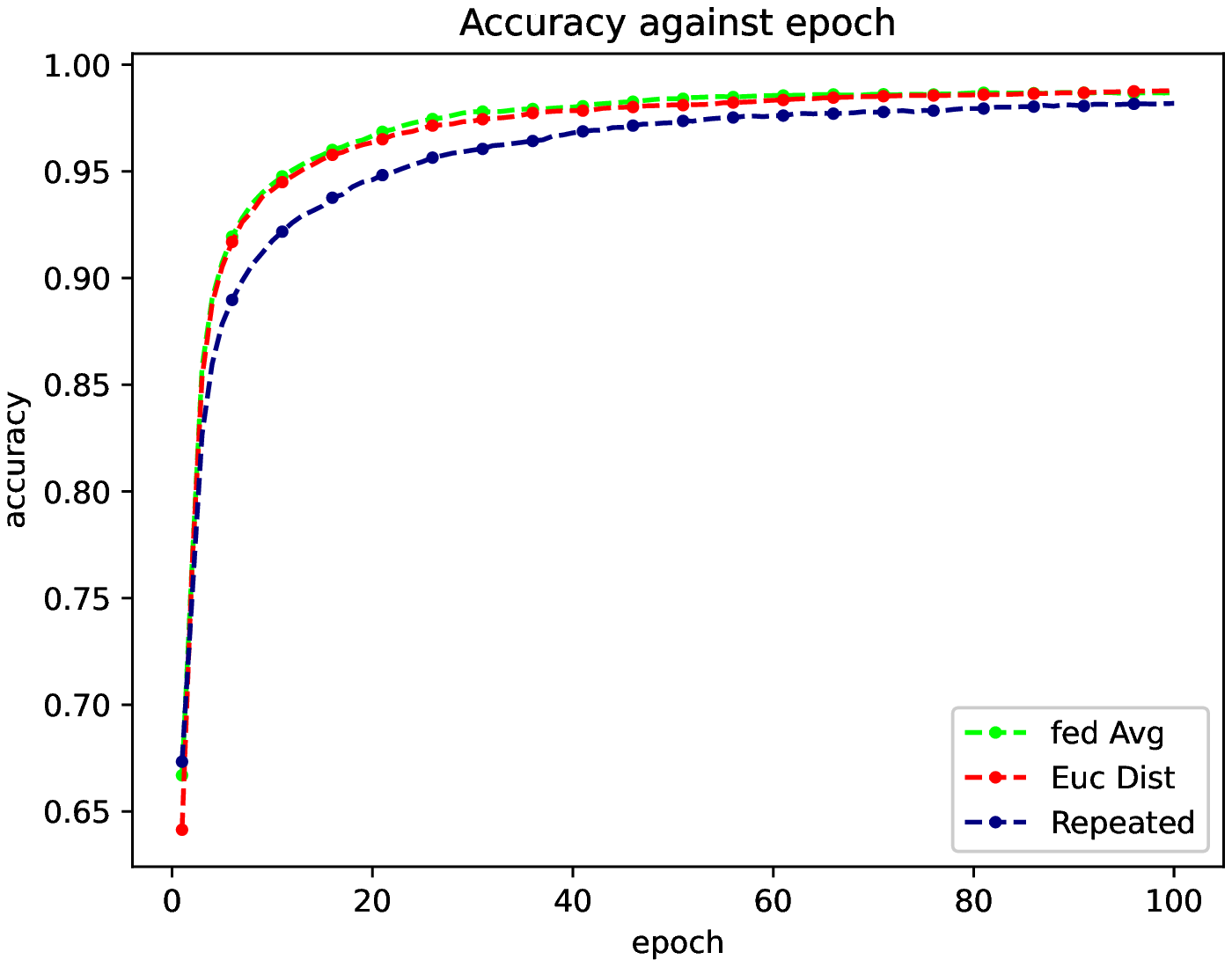}\\
  \caption{Results for experiments with all 100 participants contributed to aggregation without any attackers}
  \label{100_ecc}
\end{figure}

First, we tested our aggregation algorithm without any label-flipping attacks. We trained three different approaches with the following settings. We ran 100 synchronization rounds with a learning rate set to 0.01. In each round of FL, each participant is supposed to train the local model for 2 epochs. And we maintained this setting as the default setting throughout our experiments. There we extracted the average accuracy, loss, and average aggregation time. Moreover, we tested two approaches,

\textbf{All local participants participated in aggregation} According to our experiments, Fig. \ref{100_ecc} shows that the proposed algorithm performs the same as the FedAvg and is slightly better than \cite{Shuhao2019}. We recorded an average accuracy of 98.78\% and an average aggregation time of 1.1467s. The results are shown in Table \ref{test_acc_100}.

\begin{figure}[t]
  \centering
  \includegraphics[width=3.4in]{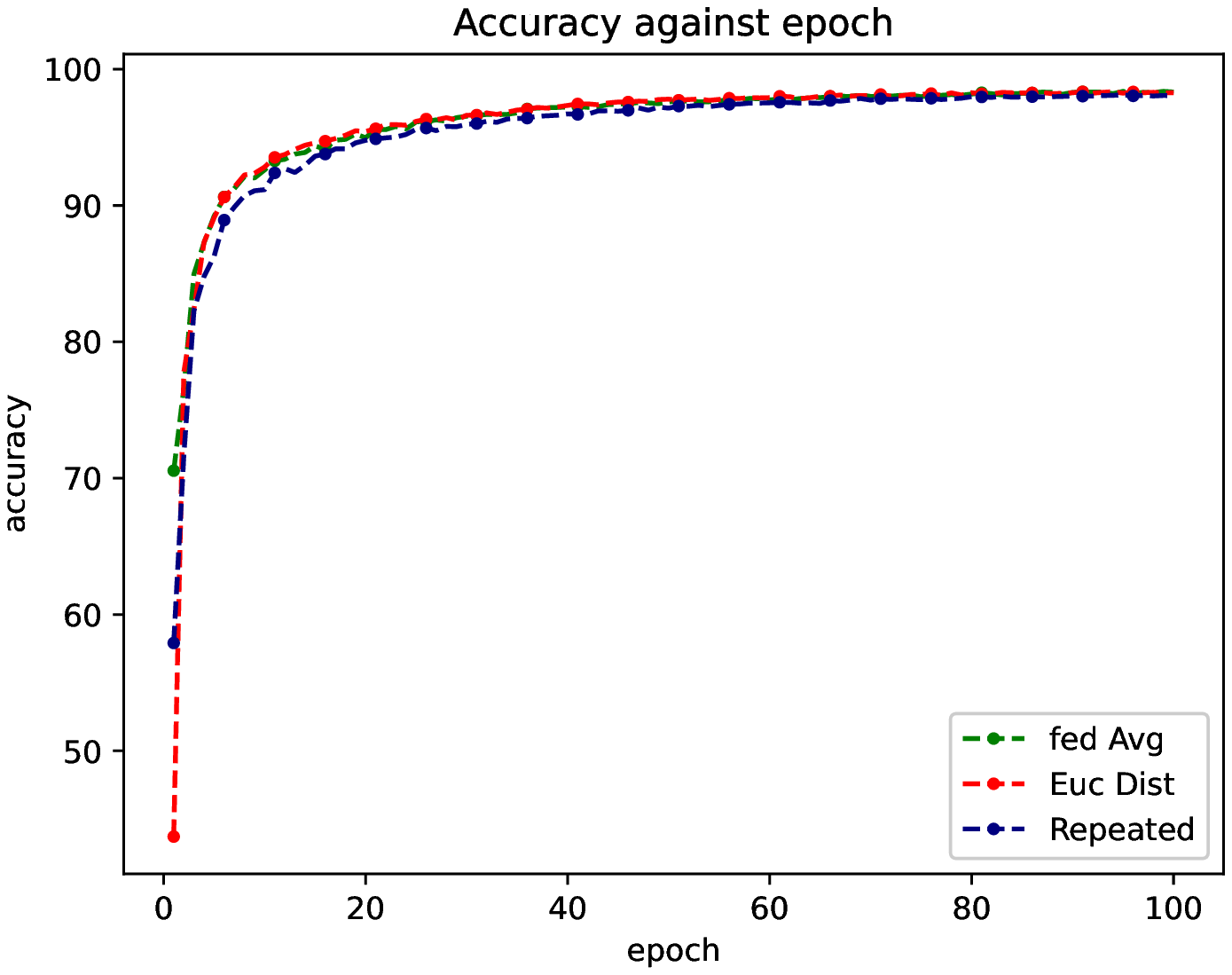}\\
  \caption{Results for experiments with 10 participants contributed to aggregation without any attackers}
  \label{10_ecc}
\end{figure}

\begin{figure}[t]
  \centering
  \includegraphics[width=3.4in]{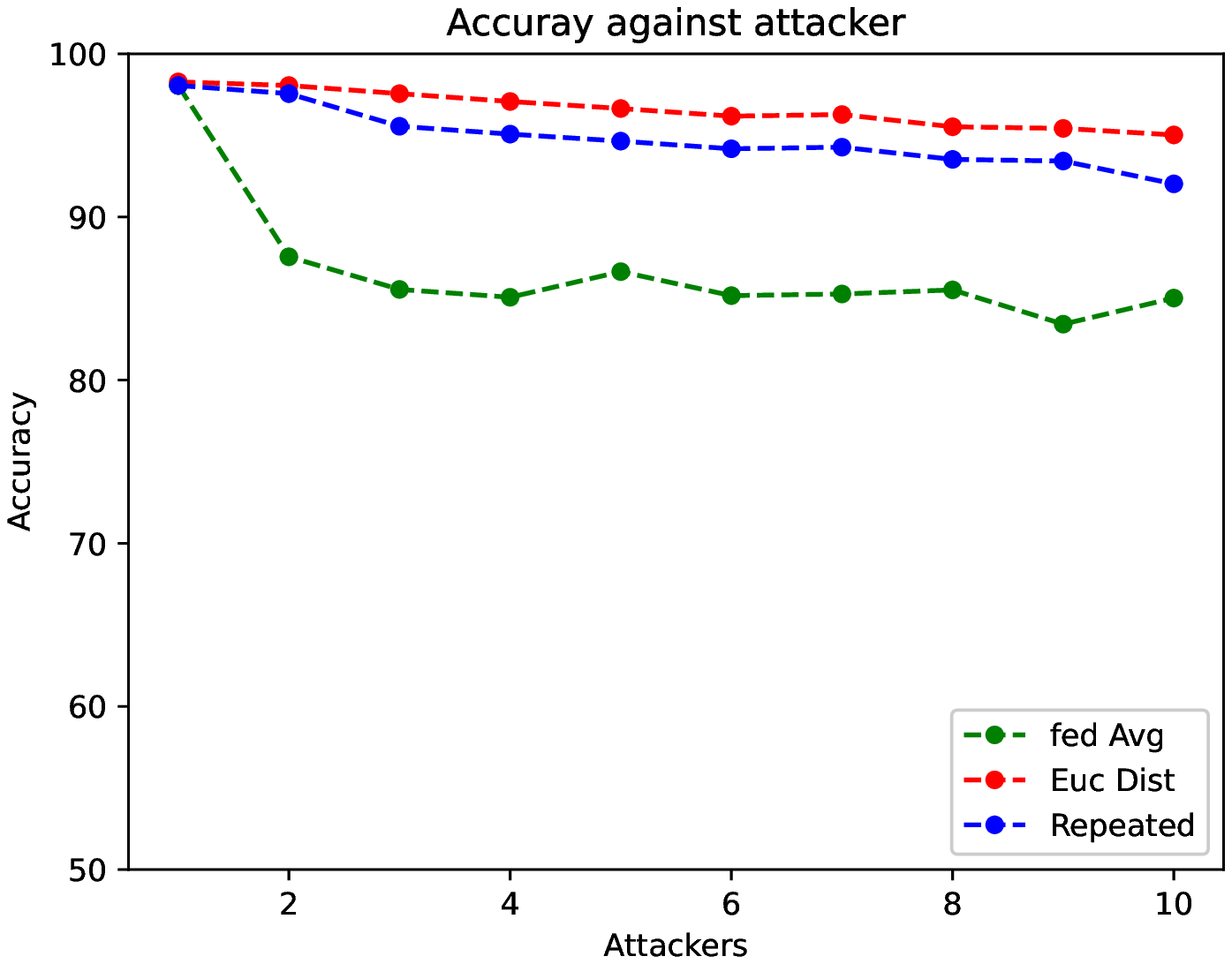}\\
  \caption{Results of label-flipping attacks with 100 users and 1-10 attacker participants in each round}
  \label{att_accuracy}
\end{figure}

\begin{table}
    \centering
    \caption{Testing accuracy without no threat attackers with all participants}
    \begin{tabular}{|c|c|c|c|}
    \hline
     \textbf{Algorithm} & \textbf{Accuracy} & \textbf{loss} & \textbf{AAT}\\
    \hline
    Euclidean Distance-based & 98.78\% & 0.0379 & 1.1467s\\
    \hline
    FedAvg & 98.68\% & 0.067 & 0.1s\\
    \hline
    Residual-based Reweighting & 98.18\% & 0.0425 & 2.5423s\\
    \hline
    \end{tabular}
    \label{test_acc_100}
\end{table}

\begin{table}
    \centering
    \caption{Testing accuracy without no threat attackers with 10 participants}
    \begin{tabular}{|c|c|c|c|}
    \hline
     \textbf{Algorithm} & \textbf{Accuracy} & \textbf{loss} & \textbf{AAT}\\
    \hline
    Euclidean Distance-based & 98.27\% & 0.0377 & 0.028s\\
    \hline
    General Averaging algorithm & 98.32\% & 0.0344 & 0.0016s\\
    \hline
    Residual-based Reweighting & 98.06\% & 0.0391 & 0.038s\\
    \hline
    \end{tabular}
    \label{test_acc_10}
\end{table}

\textbf{10 users participated for aggregation in each training round}. In this setting, as shown in Fig. \ref{10_ecc} the FedAvg algorithm performed better than both approaches. However, the difference in the accuracy is 0.05\% between the FedAvg algorithm and our approach. The results are shown in Table \ref{test_acc_10}.

\begin{table}
    \centering
    \caption{Label flipping 1-10 attacker participants, all users available, epochs-100, attack label-1, iid and 10 users per epoch}
    \begin{tabular}{|c|c|c|c|}
    \hline
     \textbf{Algorithm} & \textbf{Accuracy} & \textbf{loss} & \textbf{AAT}\\
    \hline
    Euclidean Distance-based & 95.44\% & 0.1035 & 0.028s \\
    \hline
    FedAvg & 84.68\% & 0.0389 & 0.0016s \\
    \hline
    Residual-based Reweighting & 91.44\% & 0.0489 & 0.038s \\
    \hline
    \end{tabular}
    \label{table_label_flipping_acc_attack}
\end{table}

The significant outcome of both of these experiments was that the Average aggregation time for our approach was recorded at 1.1467s while the Residual-based reweighting algorithm was recorded at 2.523s. Most importantly, our approach performed well and reduced the aggregation time from 54.55\%. 

\subsection{Results on Label-flipping Attacks}
In label-flipping attacks, attackers flip the labels of training examples in the source class to a target class and train their models accordingly. 
In the MNIST experiment, we simulate FL with 100 participants, within which 0 to 10 of them are attackers. Each participant contains images of two random digits. The attackers are chosen to be some participants with images of digit 1 and another random digit since they are flipping the label from 1 to 7. We ran 100 epochs  with a learning rate set to 0.01. In each epoch, each participant is supposed to train the local model for 2 epochs, but the attackers can train for arbitrary epochs. 

\begin{figure}
  \centering
  \includegraphics[width=3.4in]{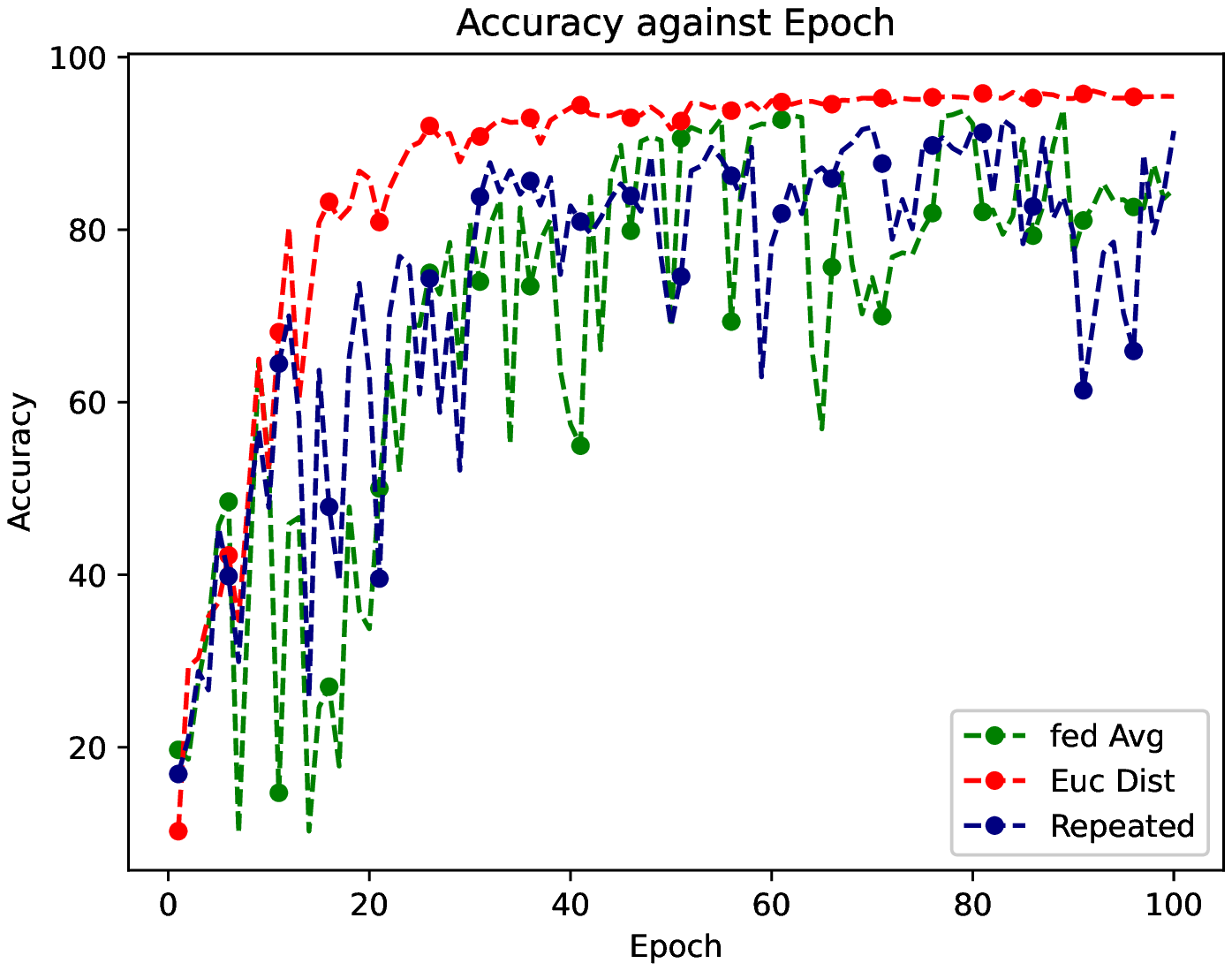}\\
  \caption{Results of label-flipping attacks with 100 users and 1-10 attacker participants in each round}
  \label{label_fliiping_attack}
\end{figure}

Fig. \ref{label_fliiping_attack} shows how accuracy differs in each epoch for the algorithms we compare. We can see that our aggregation logarithms converged between epochs 35-40 while other aggregation algorithms never fully converged. Moreover, the purposed algorithm is more stable while the other two algorithms are suffering from their unstable accuracy.

\begin{figure}
  \centering
  \includegraphics[width=3.4in]{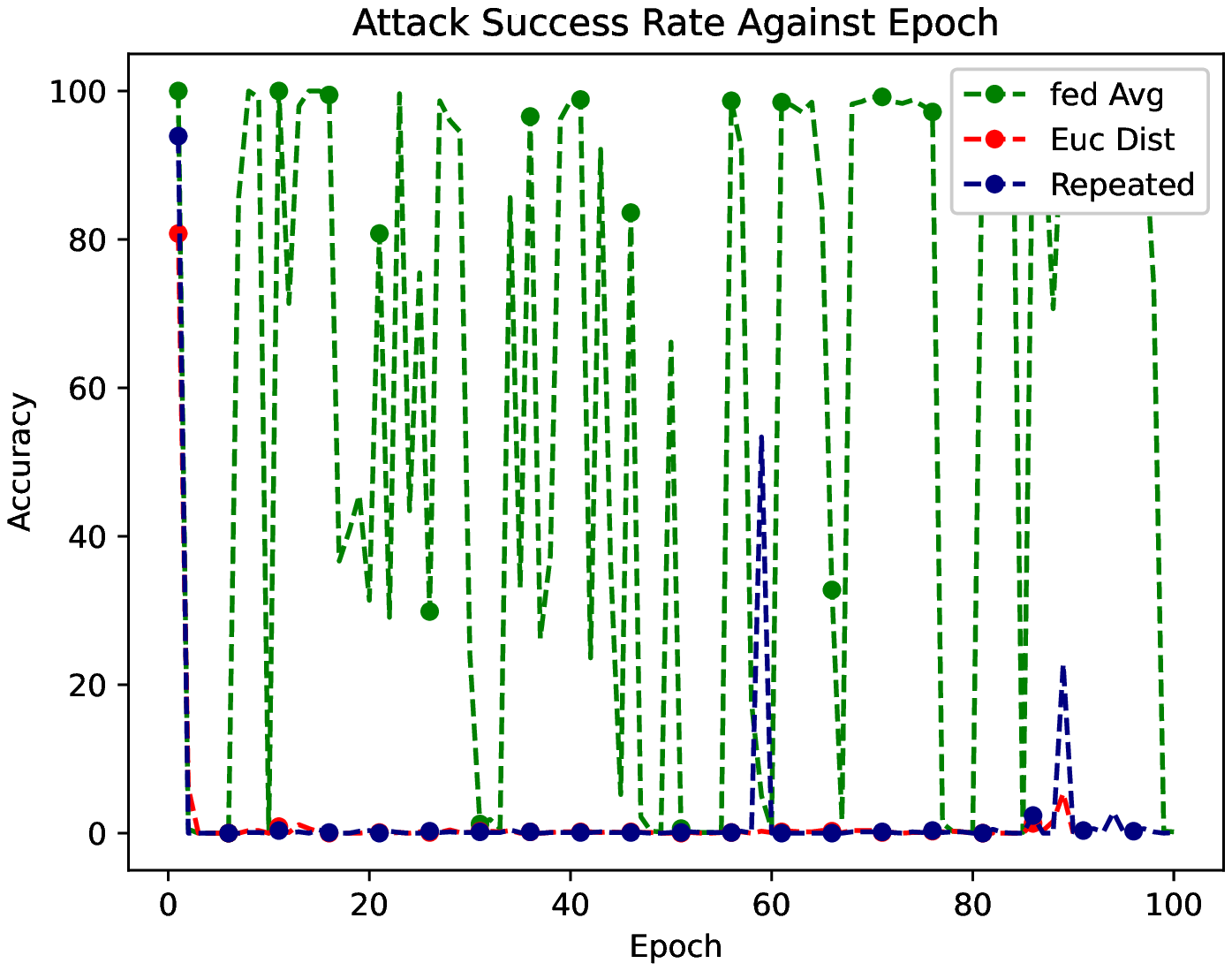}\\
  \caption{Attack success rate of label-flipping attacks with 100 users while 5 of them are malicious participants}
  \label{attacker_success_rate}
\end{figure}

We extracted the attacker success rate in each epoch under the above setting. As shown in Fig. \ref{attacker_success_rate} our approach outperformed both algorithms. However, the residual-based reweighting algorithm performed well and recorded an average attack success rate of 1.8643\% while our aggregation algorithm recorded a 1.1962\% attack success rate. Moreover as Table \ref{table_label_flipping_acc_attack} states, again our algorithm recorded less model aggregation time than the residual-based reweighting algorithm.

The results are shown in Fig. \ref{att_accuracy} where attackers train the models with 2 more epochs to enhance the attacks. Our algorithm outperformed all other methods and is robust when the number of attackers increases.

Besides, FedAvg is a coordinate-wise operation, our algorithm accumulates a weight for each model by its similarity to the most recent global model. Even though the residual-based reweighting algorithm is an effective model-wise reweighting approach as ours the time complexity is ${O(n^2)}$ while our algorithm's time complexity is ${O(n)}$ and is more efficient. 

\section{Conclusion}

Federated Learning (FL) utilizes private data on multiple devices to train a global model. However, the simple aggregation algorithm in FL is vulnerable to malicious attacks. To tackle this problem, we present a novel aggregation algorithm with Euclidean distance. Our experiments on computer vision show that our approach is more robust and efficient for label-flipping attacks than the prior aggregation methods. As future work, we are focusing to test our algorithm in different and complex datasets with backdoor poisoning attacks, vertical FL, and homomorphic encryption. We hope our proposed aggregation algorithm can make FL more practical and robust.

\balance
\bibliographystyle{IEEEtran}
\bibliography{IEEEabrv,references}

\end{document}